%% file: root.tex
\documentclass[letterpaper, 10 pt, conference]{ieeeconf}  % Comment this line out if you need a4paper

\IEEEoverridecommandlockouts                              % This command is only needed if 
\usepackage{array,multirow,graphicx}
\usepackage{hyperref}
\usepackage{amsmath}
\usepackage{caption}
\usepackage{amssymb}
\usepackage{siunitx}
\usepackage{caption}
\usepackage{subcaption}
\usepackage{tabularx}
\usepackage{url}
\usepackage{float}
\usepackage{pgf-umlsd}
\usepgflibrary{arrows} % for pgf-umlsd
\usepackage{verbatim}
\usepackage{xcolor}
\usepackage{hyperref}
\usepackage{footmisc}
\usepackage{multirow}
\usepackage{xspace}
\def\argmax{\mathop{\rm argmax}}
\def\argmin{\mathop{\rm argmin}}

\title{\LARGE \bf
Communicating human intent to a robotic companion by multi-type gesture sentences%Defining human intents for context-aware robot control using multi-type gesture sentences as a reactive system
}

\author{
Petr Vanc$^{1}$
\and
Jan Kristof Behrens$^{1}$
\and 
Karla Stepanova$^{1}$
\and
Vaclav Hlavac$^{1}$
\thanks{$^{1}$Czech Technical University in Prague, Czech Institute of Informatics, Robotics, and Cybernetics, \texttt{petr.vanc@cvut.cz}, \texttt{jan.kristof.behrens@cvut.cz}, \texttt{karla.stepanova@cvut.cz}, \texttt{vaclav.hlavac@cvut.cz}}
}

\begin{document}

\maketitle
\thispagestyle{empty}
\pagestyle{empty}

%%%%%%%%%%%%%%%%%%%%%%%%%%%%%%%%%%%%%%%%%%%%%%%%%%%%%%%%%%%%%%%%%%%%%%%%%%%%%%%%
\begin{abstract}
Human-Robot collaboration in home and industrial workspaces is on the rise. However, the communication between robots and humans is a bottleneck. Although people use a combination of different types of gestures to complement speech, only a few robotic systems utilize gestures for communication. In this paper, we propose a gesture pseudo-language and show how multiple types of gestures can be combined to express human intent to a robot (i.e., expressing both the desired action and its parameters - e.g., pointing to an object and showing that the object should be emptied into a bowl). The demonstrated gestures and the perceived table-top scene (object poses detected by CosyPose) are processed in real-time) to extract the human's intent. We utilize behavior trees to generate reactive robot behavior that handles various possible states of the world (e.g., a drawer has to be opened before an object is placed into it) and recovers from errors (e.g., when the scene changes). Furthermore, our system enables switching between direct teleoperation of the end-effector and high-level operation using the proposed gesture sentences. The system is evaluated on increasingly complex tasks using a real 7-DoF Franka Emika Panda manipulator. Controlling the robot via action gestures lowered the execution time by up to 60\%, compared to direct teleoperation.
% Human-Robot collaboration in home and industrial workspaces is on the rise. However, the communication between robots and humans is a bottleneck.
%New means of communication are needed to complement human speech and to catch up in terms of reliability and simplicity of giving orders. 
%Teleoperating robots is a key technology for humans to communicate with robots (e.g., for teaching new tasks, or remote acting in harsh environments). Say that gestures are used by humans to disambiguate language. In this paper, we propose a system that combines gesture sequences with the scene context in the form of 6D object poses to control robotic manipulations.

%Gesture language can be one of the ancillary ways of communication with the robot that lacks deeper explorations to enable gestures potential and way of communication.
%In this work, we construct a system that can process our defined gesture pseudo-language. Based on the understanding, it can guide the user to complete the specified tasks in a given environment, for example, manipulation tasks, while being able to recover from errors, when the scene changes. Controlling the robot via action gestures lowered the execution time by up to 60\%, than using a direct teleoperation method.

\end{abstract}

\maketitle

\section{Introduction}
\input{01_introduction}

\section{Related work}
\input{02_related_work}

\section{Problem formulation}
\input{03_problem}

\section{Materials and Methods}
\input{04_methods}

\section{Experimental setup}

\input{05_experimental_setup}

\section{Experimental results}
\input{06_results}

\section{Conclusion and discussion}
\input{07_conclusion}

\section{Acknowledgment}
This work was supported by the European Regional Development Fund under project Robotics for Industry 4.0 (reg. no. $CZ.02.1.01/0.0/0.0/15\_003/0000470$), MPO TRIO project num. FV40319, and   by the Czech Science Foundation (project no. GA21-31000S). P.V. by CTU Student Grant Agency (reg. no. SGS23/138/OHK3-027/23).

\bibliographystyle{IEEEtran}
\bibliography{root}

%\appendix

\end{document}

%% file: 01_introduction.tex
There are many ways how to control a robot with hand gestures. Commonly used are teleoperation \cite{Dragan_Srinivasa} and pointing (deictic) gestures \cite{10.3115/991365.991471}. 
Direct teleoperation enables users to control robots in real-time and is useful to operate the robots in places that are inaccessible or dangerous for humans (e.g., robotic surgery and safe and rescue operations in harsh environments). However, it requires the user's full attention and becomes infeasible in situations with high communication delays. Space robotics is a prime example where communication delays make some autonomy necessary during teleoperation (ground-in-the-loop) \cite{Nesnas_Fesq_Volpe_2021}.
%\deleted{Direct teleoperation enabled users full control in real-time and was useful to operate the robot in places that are inaccessible or dangerous for humans. Space robotics is a prime example where communication delays make some level of autonomy necessary during teleoperation (they call it ground-in-the-loop) \cite{Nesnas_Fesq_Volpe_2021}.}

In this paper, we propose a system that enables the user to communicate its high-level intent to the robot via gestures. Knowing the user's intent enables the robot to execute many of the required actions autonomously. Both the robot and the human can request to change the control mode. In this way, the system can disambiguate the instructions or acquire further specifications.
Different modalities allow for communicating different concepts more efficiently and complement each other. While language might be good for establishing order, gestures are more convenient for referring to objects and places, to express angle-related parameters or trajectory assignments. In general, gestures have a more limited vocabulary than speech. Therefore, one gesture might not be enough to express a complex user intent, and several gestures must be combined. In this paper, we specify gesture sentence that enables expression of a complex user intent to the robot via gesture pseudo-language. %\ks{if used my text, skip this: } 
%\deleted{Furthermore, gestures have a more limited vocabulary than speech; they also have higher uncertainty when we want to define a task, assuming that spoken words can be appropriately recognized. Therefore more gestures are needed when we want to specify an intention than by using speech. However, few measurements are more convenient to express by hand gestures, for example, the angle-related parameters or trajectory assignments. Generally, gestures are good at referring to objects and places.} 

\begin{figure}
  \centering
  \includegraphics[width=0.98\linewidth]{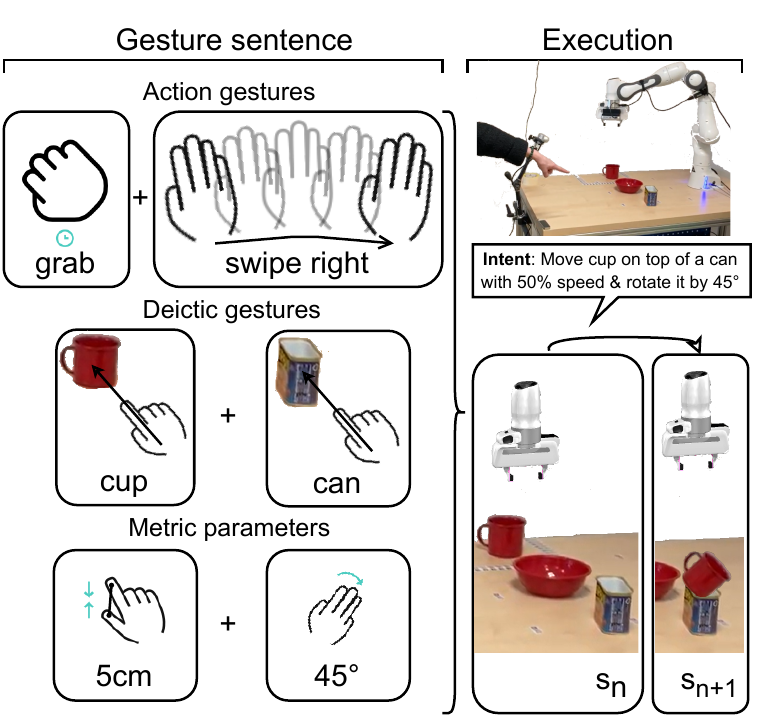}
  \caption{Human operating a robot with gestures.}
  \label{fig:introphoto}
\end{figure}

%\jb{teleoperation is used as the name of the baseline method and as what is still done in the proposed system. It isn't very clear. We could call the proposed method gesture operation or semi-autonomous teleoperation.}
In this work, we present a highly expressive gesture pseudo-language. Its purpose is to increase the efficiency and comfort of teleoperation of the robots. %and enable more natural and efficient teleoperation of robots. 
We show that in the application of tabletop manipulation tasks, the operator spends less time giving instructions compared to other systems. The main contributions of the paper are:
\begin{itemize}
    \item Definition of a gesture sentence composed of different gesture types (i.e., action, deictic, and metric gestures). 
    \item Extension of the gesture framework~\cite{Vanc2023} to handle various types of gestures, disambiguate between different gesture meanings, and determine user intent based on the observed gesture sentence.%}\deleted{System that can disambiguate information between different gesture meanings, recover from failed tasks using behavior trees and manage different (static \& dynamic) gesture sets using our teleoperation gesture toolbox}
    \item Evaluation on the increasingly complex tasks in simulation and on the real setup.%\deleted{, proving the usability of the system}.
\end{itemize}

The source code, accompanying video, and additional materials are available on the project webpage \href{http://imitrob.ciirc.cvut.cz/gestureSentence.html}{http://imitrob.ciirc.cvut.cz/gestureSentence.html}. 

%% file: 02_related_work.tex
The current systems that utilize gestures to communicate with a human mainly focus on the teleoperation of the robot (e.g., \cite{zhang2019gesture}) or use gestures to command the robot without the option to connect the human gestures to the current state of the scene (e.g., \cite{neto2019gesture}, \cite{nuzzi2021meguru}, \cite{mazhar2019real}). Zhang et al.~\cite{zhang2019gesture} uses similarly to us Leap motion sensor to detect hand movements and classify individual gestures. A basic set of gestures is used to teleoperate the robotic arm without connection to the given scene. In \cite{neto2019gesture}, wearable sensors were used to detect both dynamic and static gestures and users were provided with visual feedback. The Meguru system proposed in \cite{nuzzi2021meguru} enables users to combine several actions and connect gestures to higher-level actions. However, none of these systems enables users to specify flexibly objects to manipulate (e.g., by pointing) nor the parameters of the performed action (e.g., amount of degrees to rotate). Furthermore, these works do not enable flexible switching between direct teleoperation of the robot and gesture-based operation during execution.

In other works, such as~\cite{behrens2019specifying}, gestures are used to point to objects that should be manipulated.
In general, gestures are used in current works as one-word commands and not as real sentences containing verbs, subjects and further parameters. This means that they cannot properly pass the human intent to the robotic system.

There are several works focused on human intent inference~\cite{losey2018review}. In~\cite{8593766}, they used recursive Bayesian Inference, which can fuse multiple observations, formulated a mathematical model, and estimated human intent based on the human’s
joystick inputs. The estimated intent then influenced the amount of shared autonomy between a human and a robot. Shared autonomy methods were also used in \cite{jonnavittula2022communicating}, where the robot actively reveals its chosen convention to speed up the novice introduction to the system. Their assist approach improves the success rate of a task based on observations from joystick input than without assistance. In these works, the intent is estimated based on joystick operation. The more natural way of communication, such as human gestures and the context of the current scene, is not taken into account. In our approach, we merge more input modalities (e.g., gestures and scene information). 

%Previous works focused on natural human-robot communication focus on fusion of data from multiple modalities such as speech, gestures and that aim to merge more types of inputs while having gestures as one of the input types were merging pointing and trajectory points with speech sentences \cite{Du2018Online}. Their teleoperation approach was tested on trajectory execution for soldering.

In our previous work~\cite{Vanc2023} we showed how to estimate human intent from a combination of the scene and gesture vector. However, this concept wasn't tested in a real setup and only one type of gesture was used at a time with a simulated region of interest.
In this paper, we propose a more general gesture pseudo-language that uses a sequence of gestures of various types (i.e., static, dynamic, deictic, teleoperation) to define human intent. The system is evaluated on a set of tasks with increasing complexity on a real robot.

%To obtain the autonomous behavior of an agent, behaviour trees \cite{Colledanchise_Ogren_2018} was used to deal with robotics tasks, where any part may be prone to failure, when using the behavior tree, the robot can easily recover from changes. There may not be a work that would combine human intent execution while implementing a behavior tree approach. %\cite{https://doi.org/10.48550/arxiv.2106.01650}

%% file: 03_problem.tex
\label{sec:problem3deictic}
\textit{Problem definition}: Given an observation $\textbf{o}$, compute human intent $\textbf{i}$ and based on it generate a sequence of robotic actions $\textbf{a}$ to fulfill this intent.

\textit{Observation tuple} defined as $\textbf{o} = [\textbf{h}, \textbf{p}, \textbf{s}]$ is composed of time-series of hand gesture features $\textbf{h}$ and deictic gesture features $\textbf{p}$, both of them are computed from the raw hand data. $\textbf{s}$ represents the context of the situation and state of the system. It includes the configuration of the objects and the robot (robot end-effector, object positions, etc.).

%The \textit{Observation tuple} $\textbf{o}$ is composed of the scene state \jb{add variable for scene state} and the detections of individual human gestures $\textbf{G}$. 
The \textit{human intent} $\textbf{i}$ represents the desired change of the world state (e.g., how objects on the scene should move, if the robot gripper should open, etc.). To be able to estimate the human intent from the observation $\textbf{o}$, we have to know (or learn) also the mapping  $\mathcal{M}\colon \mathbf{i} = \mathcal{M}(\mathbf{o})$ %\jb{consider: $\mathcal{M}\colon \mathbf{o} \mapsto \mathbf{i}$}. 
The intent is mapped to the robot actions by mapping $\mathcal{A}\colon$ $\mathbf{a} =\mathcal{A}(\mathbf{i})$.

\begin{figure}
  \centering
  \includegraphics[width=0.6\linewidth]{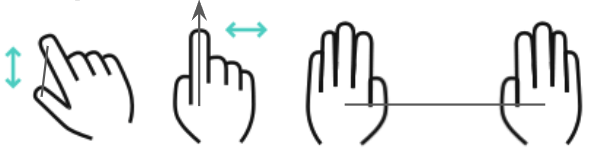}
  \caption{Metric gesture example, they are auxiliary and optional. Pinch distance (left), point direction (center), hands distance (right)}
  \label{fig:metricexample}
\end{figure}

%\jb{The following paragraph reintroduces $\mathbf{o}$. It should be merged.} %First, gesture classification task $\mathcal{G}$ has to be solved.
Individual gestures can signal different types of information (e.g., desired action, parameters of the action, or the target object). Our system fuses several gesture demonstrations into a gesture sentence with a subject, verb, objects, and further quantitative and qualitative parameters. Here we list the main types of gestures. More information on individual gestures can be found in the Sec.~\ref{sec:methods}:
\begin{enumerate}
    \item Action gestures (static and dynamic): individual gestures are mapped to various intended actions, such as pick up, open, or push.
    Based on the pre-trained set of gestures $\textbf{G}$, hand observations $\textbf{h}$, and classifier $\mathcal{G}$, our system outputs probabilities of individual gestures \textbf{g},
    $$\textbf{g} = \mathcal{G}(\textbf{h}, \textbf{G}).$$ 
    \label{sec:problem3deictic}
    \item Deictic gesture. This gesture enables the user to select region of interest (objects) in the scene. The direction vector $\textbf{p}_{\mathsf{line}}$ (direction of the pointing finger or hand palm direction) is extracted from the tracked hand configuration. Considering known object poses in the scene  $\textbf{o}_\mathsf{poses}$, probabilities of individual objects to be the target object  can be estimated based on $\textbf{to}_\mathsf{dists}$. Vector $\textbf{to}_\mathsf{dists}$ represents the distance of each object from $\mathcal{O}$ to the direction vector:
    %This method needs object positions in the scene. When turned on, with this gesture, you can move attention to a given object in the scene.
    $$\textbf{to}_\mathsf{dists} = \mathcal{D}(\textbf{p}_\mathsf{line}, \textbf{o}_\mathsf{poses})$$
    Individual distances are computed by finding the closest distance from the object $s_i\in\mathcal{O}$ to the direction line:
    $$d_i^2 = \frac{ | (\textbf{p}_2-\textbf{p}_1) \times (\textbf{p}_1 - \textbf{s}) |^2 }{ | \textbf{p}_2 - \textbf{p}_1 |^2 },$$
    where $\textbf{p}_1$ and $\textbf{p}_2$ are two random points from the 3D line and $\textbf{s}_i$ is a position of the object $i$, all vectors are 3D Cartesian vectors.
    \item Metric information. These gestures are optional and enhance the input gesture information (otherwise default values are used). They are continuous variables, e.g., distance between index finger and thumb is mapped to the parameter value such as distance of the push action or the angle when picking up an object (see examples in Fig.~\ref{fig:metricexample}).
     \item Direct teleoperation of the robot. Some action gestures might engage teleoperation mode. When active, the tracked hand center palm position and rotation around $z$ axis is directly mapped onto the robot position and 7th DoF rotation, respectively.
\end{enumerate}

Finally, gesture sentence $S$ can be defined as a tuple of observed action gestures, objects specified by deictic gestures, and metric parameters specified by metric gestures:
\begin{equation}
\label{eq:sentence}
\textbf{S} =  (\textbf{action}, object(s), \textit{metric parameters}).
\end{equation}
While the \textbf{action} is always required, object specification is required only for specific actions and metric parameters are auxiliary (if not defined, default values are used).

Individual sentences corresponds to the user intent $\textbf{i}$ (as introduced in~\cite{Vanc2023}), i.e. target action $\textit{ta}$, target object $\textit{to}$, target location, and other auxiliary metric parameters ${\mathbf{ap}}$: 
%Finally, user intent can be defined as a tuple of target action $\textit{ta}$, target object $\textit{to}$, and target auxiliary metric parameters ${\mathbf{ap}}$. 

\begin{equation}
\label{eq:intent}
    \textbf{S}  \sim \textbf{i} = (\underbrace{\underset{\mathcal{O}}{\argmax}(\textbf{i}_{ta})}_{ta}, \,\underbrace{\underset{\mathcal{O}}{\argmin}(\textbf{to}_\mathsf{dists})}_{to}, \, \mathbf{ap}),
\end{equation}

where $\textbf{i}_{ta} = \mathcal{M}_a(\textbf{g})$ is \textit{target action intent} vector probabilities generated from gestures vector $\textbf{g}$, $\textbf{to}_\mathsf{dists} = \mathcal{D}_a(\textbf{p}_\mathsf{line}, \textbf{o}_\mathsf{poses})$ are \textit{target object} vector probabilities. $ta$ represents \textit{target action intent}, $to$ \textit{target object}, and $\mathbf{ap}$ set of auxiliary parameters.

An example of gesture sentence is:
\begin{equation}
\label{eq:example_sentence}
S = (\textbf{thumbsup}, \textit{[mug, bowl], [5cm]}).
\end{equation}

This corresponds to the user intent (\textit{"Move mug to the bowl by 50\% speed"}):
\begin{equation}
\label{eq:example_intent}
i = (\textbf{move}, \textit{mug, [bowl, 50\%]})
\end{equation}

% \begin{equation}
% \label{eq:intent}
% \textbf{S} = \textbf{i} = (ta, to, \mathbf{ap})
% \end{equation}

The \textit{robotic action sequence} $\textbf{a}$ is generated from estimated human intent $\textbf{i}$. In this work, it is solved by behavior trees \cite{Colledanchise_Ogren_2018} (see example in Fig.~\ref{fig:BTreeDiagram}). The whole structure can be seen in Fig.~\ref{fig:CVWW22_systemdiagram}. 

\subsection{Complexity of the gesture sentence}
\label{sec:complexity}

The gesture sentences have variable complexity based on the number of variables that have to be specified for the given robotic action. The gesture sentence has only one required parameter ($ta$).
We define the complexity of the gesture sentence $S = (ta, \textbf{to}, \mathbf{ap})$ corresponding to robotic action $a$ with $N$ parameters $\mathbf{p}$ ($a(\mathbf{p})$) as:
$$S_c = N = len(\mathbf{to})+len(\mathbf{ap})$$.
 Simple robotic actions need to specify only the type of the action and have no additional parameters nor object dependency (e.g., move right based on grid world step). In this case, gesture sentence complexity $S_c=0$. For actions such as \emph{Pick up a cup}, we must specify the action and the target object ($S_c=1$). Complex robotic actions can depend on two or more scene objects (e.g., swap the position of two objects).
%(corresponding to increasingly complex robotic actions). These actions differ in the amount of information that has to be defined by the gestures (i.e. robotic action parameters). 

Our system contains the following set of robotic actions with increasing number of parameters:

\begin{itemize}
    \item 0-parameters: Cartesian moves, rotate, place
    \item 1-parameter: Pick up, Put, Pour, Open, Close
    \item 2-parameters: Put, Pour, Swap
\end{itemize}

Note that e.g. action \emph{Put} can either have 1- parameter (if we are holding the target object then only action and target location has to be specified), or 2- parameters (if the gripper is empty we have to specify action, target object and target location).

%% file: 04_methods.tex
\label{sec:methods}
The problem described in Section 3 can be divided into several parts: user hand detection and bone reconstruction, gesture detection, user intent recognition, and robotic action generation and its execution by the robot. In our system, we don't consider hand detection and bone reconstruction. The overall system is visualized in Fig.~\ref{fig:CVWW22_systemdiagram}.

 Our system considers 1) Gesture recognition $\mathcal{G}$, 2) Intent classification $\mathcal{M}$, and 3) Robotics action generation~$\mathcal{A}$. Parts of the system are introduced in more detail in the next sections and the last subsection connects all methods together (Sec.~\ref{sec:methodsinteraction}).

\subsection{Real-time gesture recognition}

We differentiate between static and dynamic gestures. Static gestures contain only a single time frame while typically having more features and dynamic have more time frames and have fewer features. Features are extracted from hand bone structure. We utilize the Leap Motion sensor \cite{Weichert_Bachmann_Rudak_Fisseler_2013} to obtain hand feature data. We utilize Gesture Toolbox~\cite{gesturetoolbox} to manage gestures for a given session. The set of gestures may be adapted to current needs.

\textit{Static gestures} feature set is hand-crafted, processes hand bone structure, and has 57 features: 1) distances between fingertip positions, alongside with the palm center position, and 2) joint angles between hand bones $\alpha_1, \ldots, \alpha_n$ constructed as angles between direction vectors. Direction vectors are made from each bone's start and end positions.
%Evaluation of hand configuration is made with a given frequency by using the last detected hand frame.
%\begin{equation}
%    \textbf{h} = [\alpha_1, \ldots, \alpha_{42}, || \textbf{x}_{f1}, \textbf{x}_\mathsf{palm} ||, || \textbf{x}_{f1}, \textbf{x}_{f2} ||, || \textbf{x}_{f1}, \textbf{x}_{f5} || ]
%\end{equation}
The best-suitable method to classify static gestures has proven to be a probabilistic neural network model \cite{Emaasit_2018}. Other approaches, including convolutional neural network and deterministic (hand-picked threshold) approach, were tested while scoring worse (see Sec.~\ref{sec:gestureclassification}).

\textit{Dynamic gestures} are evaluated as hand trajectories over time. The used sample configuration is composed only of the palm center position (Cartesian) in each time step:

\begin{equation}
    \textbf{h} = \left[ \begin{vmatrix} x_1 \\ y_1 \\ z_1 \end{vmatrix},
    \begin{vmatrix} x_2 \\ y_2 \\ z_2 \end{vmatrix},
    \cdots
    \begin{vmatrix} x_N \\ y_N \\ z_N \end{vmatrix}
    \right],
    \label{dynamicobservation}
\end{equation}
Where $N$ is a number of trajectory points obtained from recording with a frequency around $90$Hz, we reduce the frequency of received frames to $f = 20$Hz without any significant loss \cite{Forbes_Fiume_2005}. We set the detection frequency to $10$Hz, each detection takes the last time window with adjustable length (here set to $1$s).

Dynamic gestures are learnt from a set of demonstrations and represented as probabilistic motion primitives \cite{Paraschos_Daniel_Peters_Neumann_2013}. As a classifier, we use the Dynamic Time Warping method (DTW) \cite{Muller_2007}, which can compare two motions regardless of speed and deformations. The comparison parameter is distance $\delta$. Every gesture has its saved representative motion $(\textbf{h}_1, \cdots, \textbf{h}_D$), where $D$ is a number of dynamic gestures in the current set, and $\textbf{h}$ is defined in \ref{dynamicobservation}. Then the most probable gesture is obtained with $\argmin((\delta)_1^{D}) = \argmin(DTW(\textbf{(h)}_1^D, \textbf{h}_s))$, where $\textbf{h}_s$ is sample motion, which we want to classify.

\subsection{Deictic gesture}

The deictic gesture is a special type of static gestures with a special methodology for detecting intended target point or object (see Sec.~\ref{sec:problem3deictic} how the closest object to the direction line is found). In the case, that we want to select a given position on the table instead of a target object (e.g., when placing an object to position), the deictic approach finds the contact of pointing 3D line with the workspace table. 

The solution (the closest object) may be computed in each frame, but its information is used only when the user points to an object or place. This gesture together with action gestures enables us to detect the desired human intent. The accuracy and usability of this gesture is evaluated in a separate experiment in Sec.~\ref{sec:Deictic_gestures_evaluation}.

\subsection{Direct robot teleoperation}

Direct teleoperation, when enabled, maps the chosen point in the user's hand (e.g., the center of hand palm) to the robot end-effector in real-time. To achieve smooth control, we update new values to the robot with a frequency of at least $10$Hz. We use position control wrapped around velocity control to avoid shaking when changing the robot's course while the robot is moving.

The mapping feature vector comprises a palm Cartesian position ($p = [x,y,z]$) and hand rotation around the $z$-axis $\theta$ to accommodate the robot's 7th degree of freedom. This is important when we want to grasp an object in the wrong configuration. We tested mapping the full rotation of the human hand, but the teleoperation experience and accuracy were worse because of the user concentration on that many angles.

Direct teleoperation enables a demonstration of unknown lower-level motion primitive to teach the robot a new skill. Later it can be mapped to higher-level commands (gestures). Furthermore, it enables to adjust position of the robot when the object detections are not perfect.

%The direct teleoperation of the robot is inferior to action gesture recognition because the robot should understand the environment in which is and act based on higher-level commands. However, direct teleoperation is needed when we want to demonstrate unknown lower-level motion primitive to teach the robot a new skill. When it is known, then teleoperation is no longer needed.

\subsection{Intended action estimation}

The intended target action $\textbf{i}_{ta}$ is estimated from the hand gestures $\textbf{g}$ and context information such as scene $\textbf{s}$ same as in~\cite{Vanc2023}. This mapping task ($\mathcal{M}$) is treated as a classification problem, a shallow probabilistic neural network is sufficient for having few scene features on input, having two fully connected hidden layers. The output is a categorical distribution \cite{murphy2013machine} commonly used for getting a discrete output. The automatic differentiation variational inference (ADVI) \cite{Kucukelbir_Tran_Ranganath_Gelman_Blei_2016} fits the network weights using a Kullback-Leibler divergence loss function. We are using PyMC \cite{Emaasit_2018} framework.

\subsection{Action sequence generation and execution} 

\begin{figure}[h]
  \centering
  \includegraphics[width=0.95\linewidth]{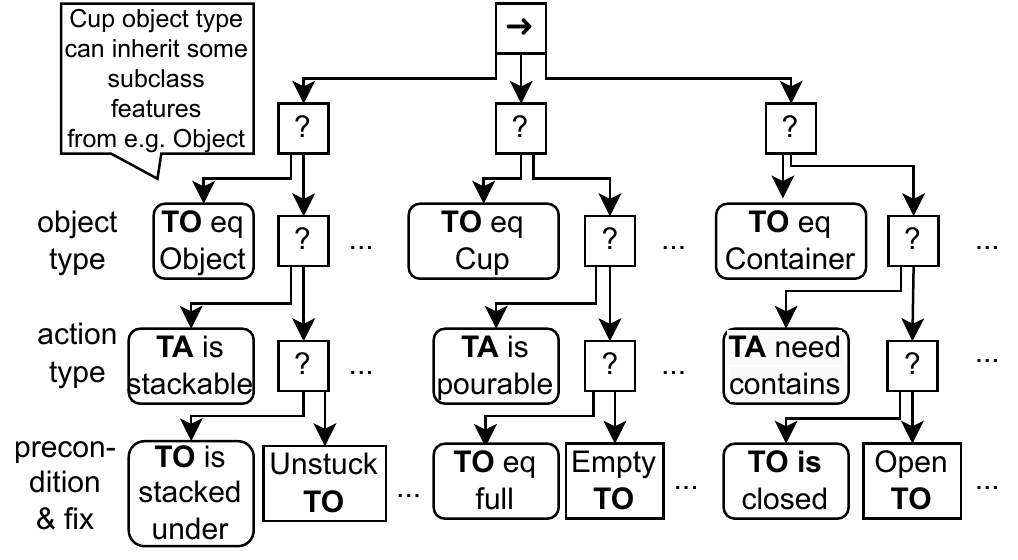}
  \caption{Behavior tree structure example is solved as completing object state preconditions. For example, picking up an object stacked up with other objects requires to first unstacking all objects above it.}
  \label{fig:BTreeDiagram}
\end{figure}

The intent $\textbf{i}$ tuple (\textit{target action}, \textit{target object}, (optionally) auxiliary parameters) is mapped to a set of robotics actions $\textbf{a}$. We utilize a behavior tree to encode this high-level policy which enables us to handle more complex tasks, for example, grasping an object that is not directly visible. The high-level policy of behavior tree recalculates all tasks for every executed action, so it can handle situations when the previous action fails. The behavior tree is embedded as generating precondition actions to fulfil the desired intent without the need to complete it manually (see Fig.~\ref{fig:BTreeDiagram} for an example of behavior tree used in our experiments).

\subsection{Human-Robot interaction}
\label{sec:methodsinteraction}

%The encapsulating interaction between a human and a robot is episodic. The user fills the sentence with gestures after all needed information is obtained, then the robot can responds. 

\begin{figure}[h]
  \centering
  \includegraphics[width=0.98\linewidth]{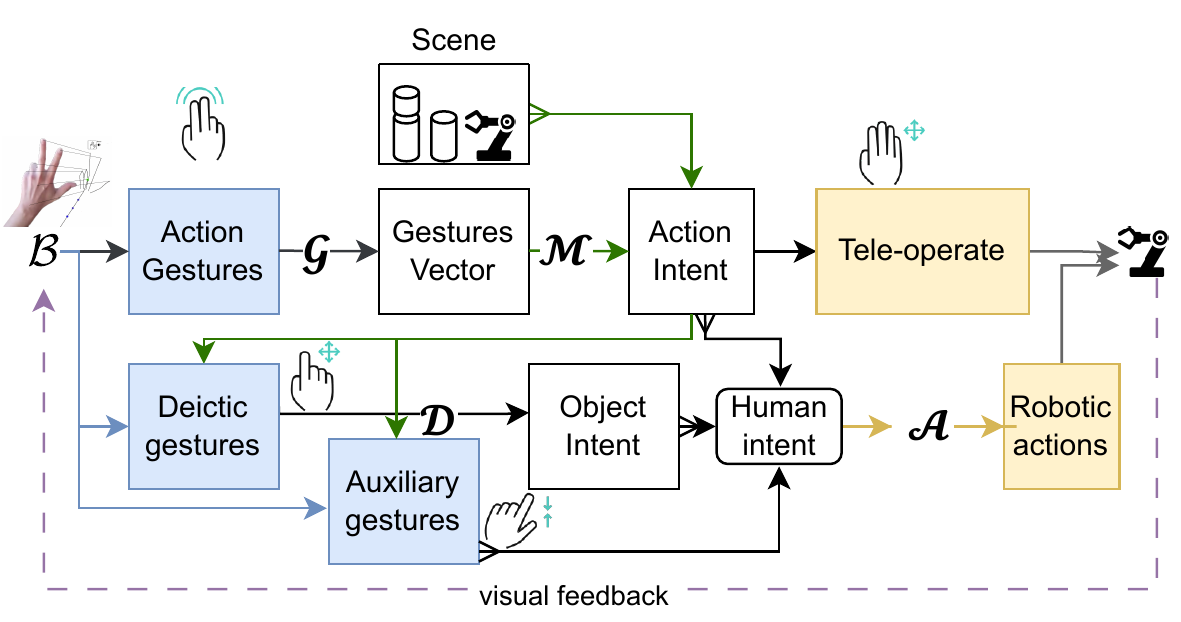}
  \caption{System diagram. $\mathcal{G}$ represents gesture classifier, $\mathcal{M}$ is mapping from gestures to human intent, $\mathcal{A}$ is robotic action generation using Behaviour tree, $\mathcal{D}$ is Deictic gesture execution. Blue blocks represent user inputs, and yellow blocks are robot output modes.
  }
  \label{fig:CVWW22_systemdiagram}
\end{figure}

The interaction between a human and a robot uses simple gesture sentences, where consecutive gestures compose a gesture sentence (see Eq.~\ref{eq:sentence}) and define the user intent (see Eq.~\ref{eq:intent}). The user is expressing his intent through gestures while getting visual feedback (e.g. detected gesture, selected object, etc.) both from the graphical user interface and directly from the robot (see the accompanying video). The diagram of human-robot interaction is shown in Fig.~\ref{fig:CVWW22_systemdiagram}. The user expresses each piece of information (i.e. action intent, object intent, auxiliary parameters) within a short interaction which we call \emph{an episode}. 

\begin{figure}[h]
  \centering
  \includegraphics[width=0.95\linewidth]{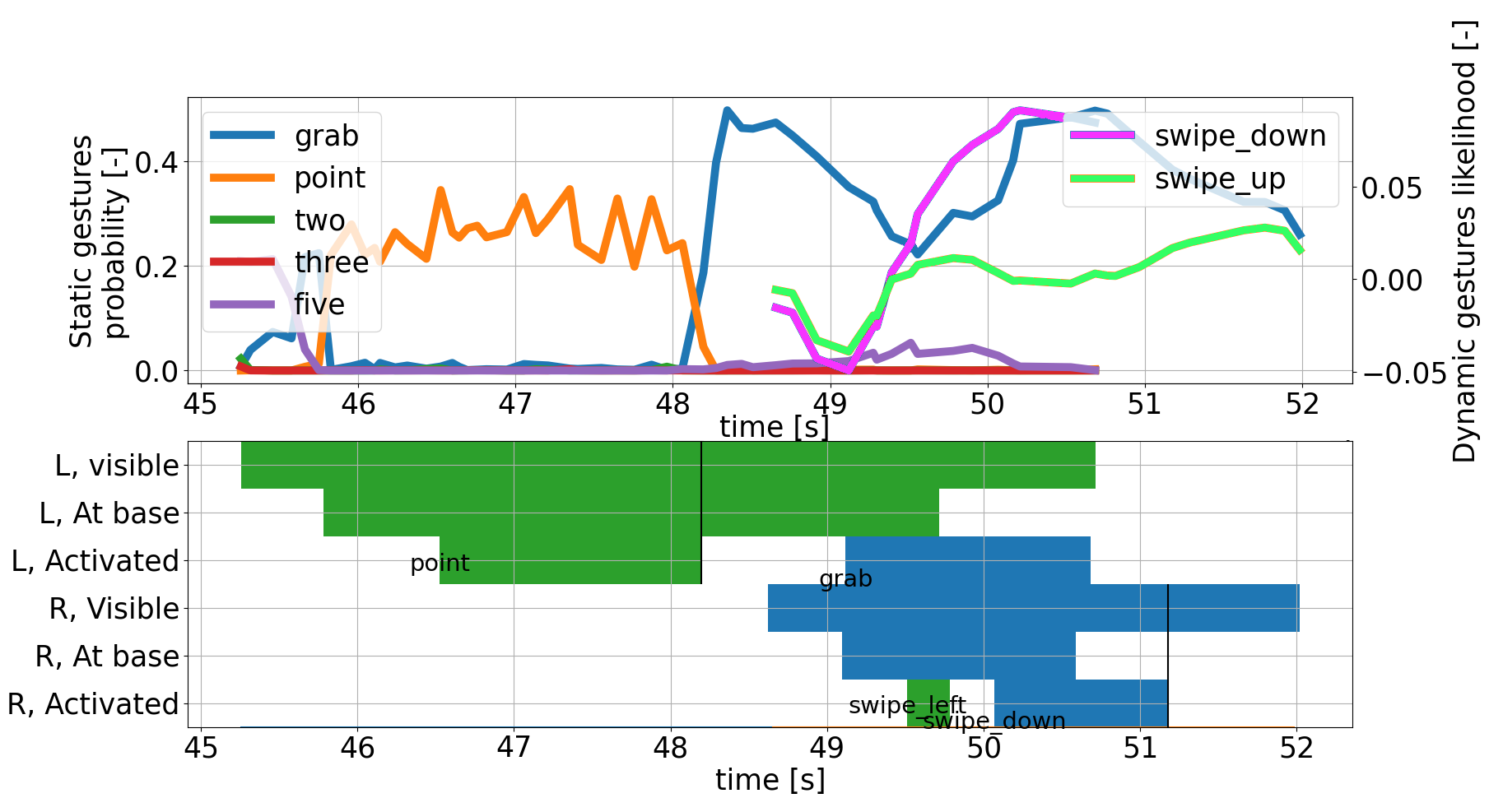}
  \caption{The gesture detection output of a single episode. The upper plot shows the likelihood of static (left legend) and dynamic (right legend) gestures. The bottom plot shows the
visibility of the hands and the activation of the gestures. First was detected the static gesture point followed by grab and dynamic gesture swipe down.}
  \label{fig:gestureobservation}
\end{figure}

We define the \textit{episode} as the time window when a human hand appears in the hand detection area and ends when it is no longer visible or the time exceeds the limit (set to $t = 3s$). In one episode, we obtain a list of triggered gestures. Static and dynamic gestures are set up as two independent threads, where we get the recognition data from each. To be signalized as detected, the gesture needs to exceed the given threshold, i.e., the amount of time when the gesture exceeds the threshold limit must be greater than $0.3s$. An example of gesture detection provided as optional feedback to the user is in Fig.~\ref{fig:gestureobservation}. 

The interaction has the following steps (see Fig.~\ref{fig:CVWW22_systemdiagram}): 1)~The user signalizes the intended action by making an action gesture, which is recognized by gesture classifier $\mathcal{G}$. The recognizer is running the whole time when the hand is visible. 2)~The action intent is estimated. Although multiple actions might be detected, in these experiments, the last detected gesture is used for determining the target action. 3)~Based on the intended action, the user is asked to provide additional information about the performed action and add the target object and auxiliary parameters (e.g. target location, distance, etc.). 4) When the sentence information is completed (or the user does not want to provide additional information), the intent tuple is filled. 5) The behavior tree of actions may be generated, and a sequence of actions is executed, leading the robot to fulfil the intended task.

%% file: 05_experimental_setup.tex
\subsection{Robot environment}

\begin{figure}
    \centering
    \begin{subfigure}[b]{0.42\linewidth}
      \centering
      \includegraphics[width=\textwidth]{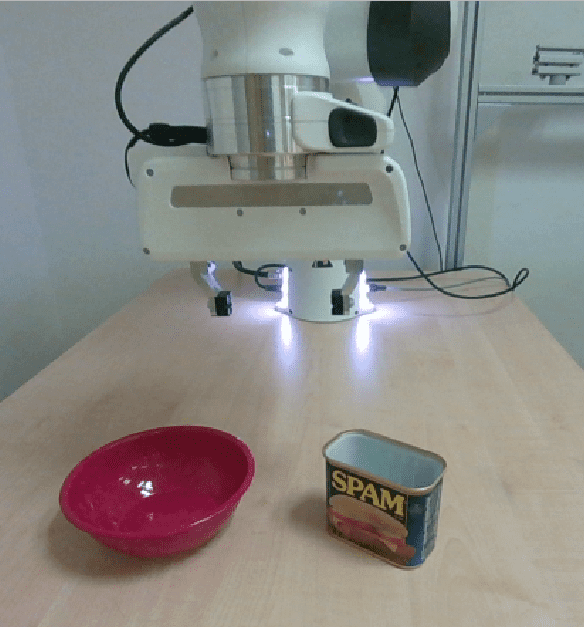}
      \caption{Real}
      \label{fig:sub1}
    \end{subfigure}%
    \hfill
    \begin{subfigure}[b]{0.41\linewidth}
      \centering
      \includegraphics[width=\textwidth,trim={0 2.4cm 0 2cm},clip]{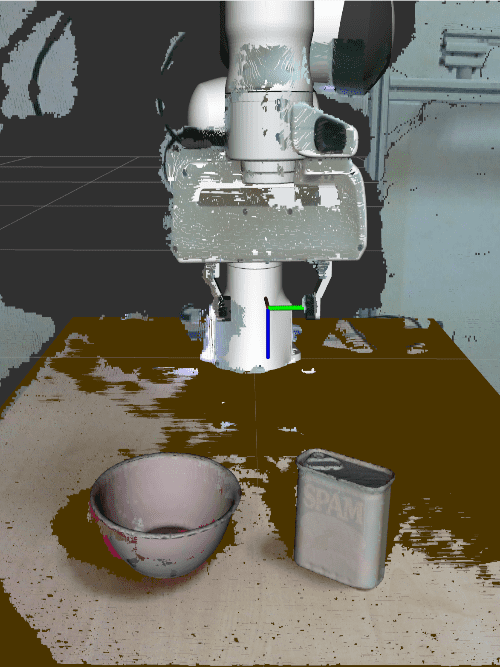}
      \caption{Belief visualization.}
      \label{fig:sub2}
    \end{subfigure}
    \caption{Cosy Pose \cite{labbe2020} object detection view using RViz visualization, which displays two detected objects. In the visualization, the meshes of detected objects (in grey) are displayed over the colored point cloud.}
    \label{fig:DeicticTestGridCoppelia}
\end{figure}

\begin{figure}
  \centering
  \includegraphics[width=0.8\linewidth]{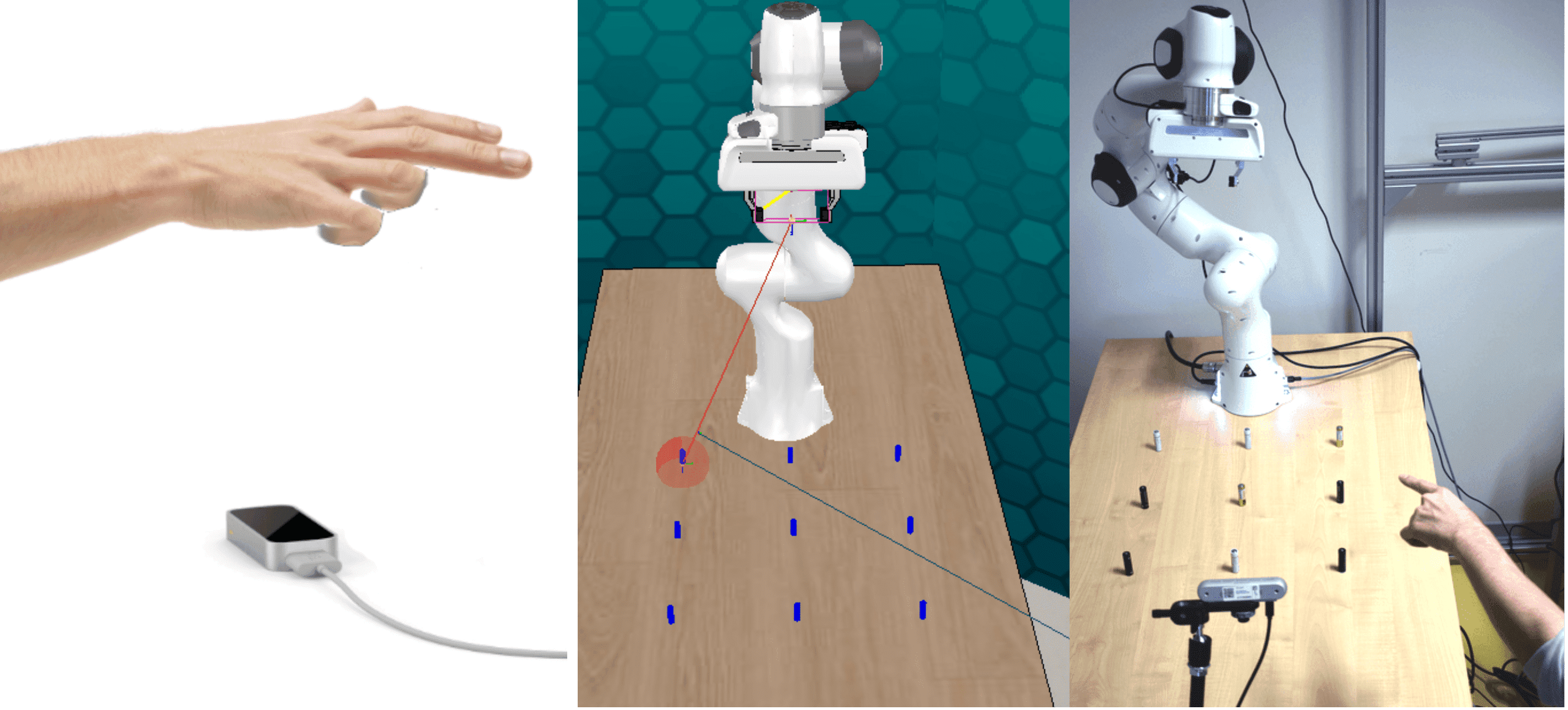}
  \caption{Experimental setup. Leap Motion Controller (left), Simulation scene in Coppelia Sim (center), Real scene (right).}
  \label{fig:experimentalsetup}
\end{figure}

Our system is tested in simulated and real environments (see Fig.~\ref{fig:DeicticTestGridCoppelia} and Fig.~\ref{fig:experimentalsetup}). For the simulator, we use Coppelia Sim \cite{Rohmer_Singh_Freese_2013} with PyRep extension tool \cite{James_Freese_Davison_2019}. The scene contains a Franka Emika Panda with 7 DOF and several manipulation objects. An Intel Realsense D455 RGBD camera opposing the robot is used for object detection. A LeapMotion sensor is attached next to the camera on the table.

Our real-world setup  (see Fig.~\ref{fig:introphoto}) takes advantage of object pose detector CosyPose \cite{labbe2020} (see Fig.~\ref{fig:DeicticTestGridCoppelia}), this allows us to get object poses in real-time. We use the YCB dataset \cite{7254318} which contains various commonly used objects in the household. This allows us to create various scene configurations. %You can see the experimental setup in Fig.~\ref{fig:experimentalsetup}.

\subsection{Gesture classification}

Our hand tracking device includes Leap Motion Controller \cite{Weichert_Bachmann_Rudak_Fisseler_2013}, which provides us with hand bone structure in real-time. We utilize our Gesture Classification framework \cite{gesturetoolbox} to manage gesture sets, handle gesture recognition and visualization of each gesture probability, as well as a graphical interface and feedback for the user.

\label{sec:setupgesturesets}
Static and dynamic gestures used in experiments include: grab, pinch, point, two, three, four five, thumbs up, swipe up, swipe down, swipe left, and swipe right gestures. The \textit{no gesture} is used to represent no movement in case of dynamic gestures and it is excluded from the list. %For some gesture examples, see Fig.~\ref{fig:01_introphoto}.

\subsection{Description of experimental scenarios}
\label{sec:scenariodescription}

%We define robotic actions with different complexity (corresponding to increasingly complex gesture sentences). These actions differ in the amount of information that has to be defined by the gestures (i.e. robotic action parameters). Simple robotic actions have no object dependency (e.g., move right based on grid world step) and complex robotic actions can depend on two or more scene objects (e.g., swap the position of two objects).

%Our system contains the following set of actions with increasing number of parameters (0-parameters mean that only action type has to be specified):

%\begin{itemize}
%    \item 0-parameters: Cartesian moves, rotate, place
%    \item 1-parameter: Pick up, Put, Pour, Open, Close
%    \item 2-parameters: Put, Pour, Replace
%\end{itemize}

We conducted a set of typical manipulation experiments of increasing complexity of the gesture sentence (see Sec.~\ref{sec:complexity} for more details, $x$ signs the object in the gripper):
\begin{itemize}
\item $\mathbf{S_c = 0}$, $\mathbf{S=(ta,-,-)}$: $(rotate,x,-)$, $(place,x,-)$, $(move,-,-)$
\item $\mathbf{S_c = 1}$, $\mathbf{S=(ta,to,-)}$: $(pick,can)$, $(open, drawer)$, $(close, drawer)$
\item $\mathbf{S_c = 1}$, $\mathbf{S=(ta,-,loc)}$: $(pour,x,bowl)$, $(put,x,bowl)$
\item $\mathbf{S_c = 2}$, $\mathbf{S=(ta,to,loc)}$: $(pour,spam,bowl)$, $(swap,can,bowl)$, $(put,spam,drawer)$
\item $\mathbf{S_c = 3}$, $\mathbf{S=(ta,to,loc,par)}$: $(rotate,spam, ang=180^\circ)$, $(pour, can, bowl, ang=60^\circ)$
\item \textbf{Multistep tasks}: stack 3 objects, tidy up 3 objects, pour 2 objects into the bowl
\item \textbf{Infeasible tasks}: move object to occupied bowl, put object to closed drawer
\end{itemize}

We are interested in the total time of task completion, the success rate for each scenario, and the user interaction time. Execution of all these tasks is visible in the accompanying videos. For experiments with the users, we selected a subset of these tasks: 1) Put an object into the bowl, 2) Swap objects, and 3) Place the object in an object-occupied bowl, demonstrating our system's ability to first move the object out automatically. The following three assistance modes are evaluated: Tasks are tested based on three modes of assistance: 1) direct teleoperation, 2) low-level control via action gestures, and 3) high-level control via action gestures.

%We define robotic actions with different complexity. They differs based on number of object arguments. Simple robotic actions has no object dependency (e.g., move right based on grid world step) and complex robotic actions can depend on two or more scene objects (e.g., replace two objects).

%Our sample action set based on increasing level of complexity contains: 

%% file: 06_results.tex
We show results for the following experiments:

\begin{enumerate}
    \item Deictic gestures - 1) time to complete, 2) accuracy
    \item Static and dynamic gestures accuracy
    \item Scenario completion
\end{enumerate}

All the scenes and tasks are set so that the robot is able to complete tasks using objects in the scene.

\subsection{Deictic gestures evaluation}
\label{sec:Deictic_gestures_evaluation}

A rough grid was created on a working table consisting of 9 small objects (see Fig.~\ref{fig:experimentalsetup}). The grid distance between two objects is set to $d = 0.2m$. The overall accuracy in detecting the intended (pointed) target object depends on the Leap Motion hand detection accuracy and accuracy of each transformation (e.g., the sensor to base, or hand to sensor).

Firstly, we tracked the hand-pointing finger bone positions, we observed high variances across measurements as the requirement is that the pointing finger is parallel to palm direction and requires an experienced user. Therefore we used values from hand palm. 

\begin{figure}[h]
    \centering
    \begin{subfigure}{.155\textwidth}
      \centering
      \includegraphics[width=.98\linewidth]{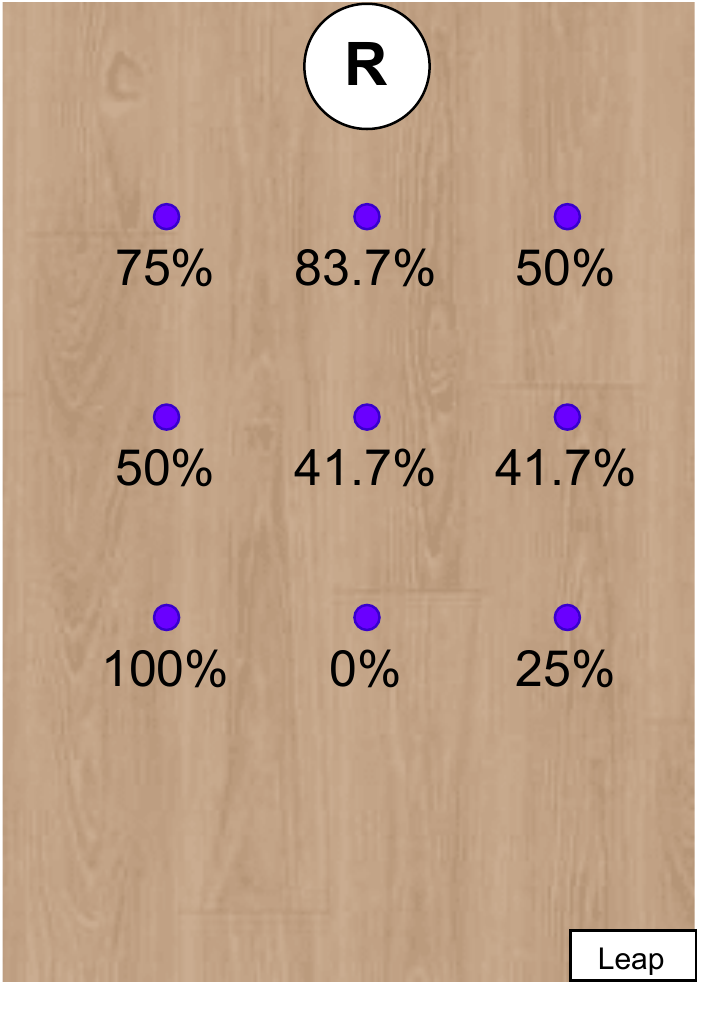}
      \caption{Naive}
      \label{fig:sub1}
    \end{subfigure}
    \begin{subfigure}{.155\textwidth}
      \centering
      \includegraphics[width=.98\linewidth,trim={0 0cm 0 0cm},clip]{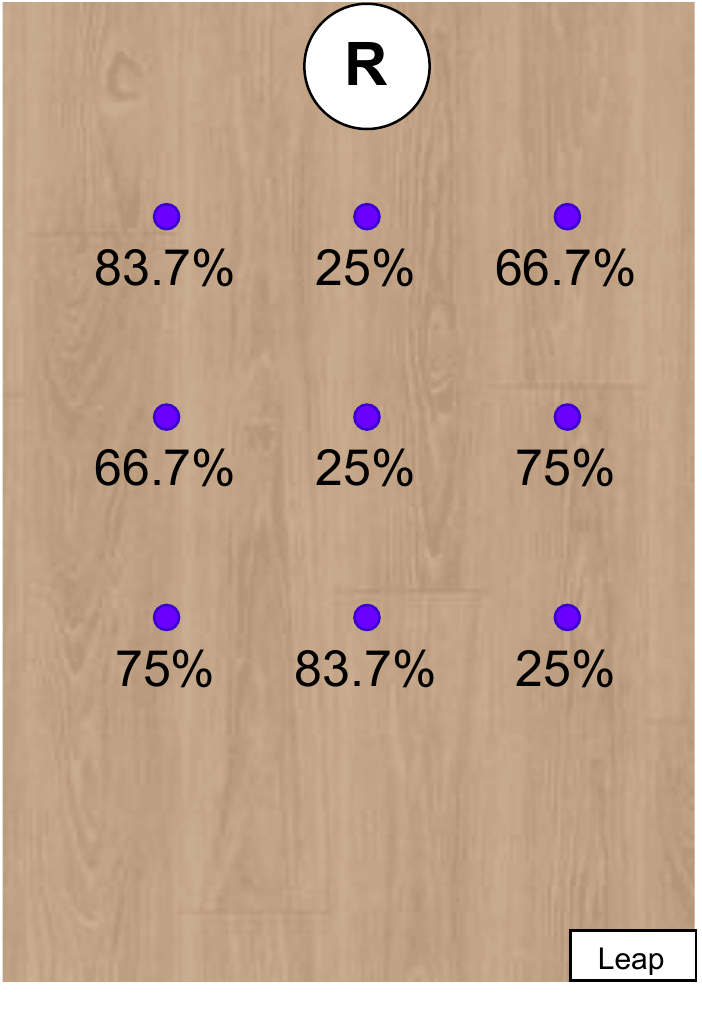}
      \caption{Trained}
      \label{fig:sub2}
    \end{subfigure}
        \begin{subfigure}{.155\textwidth}
      \centering
      \includegraphics[width=.98\linewidth,trim={0 0cm 0 0cm},clip]{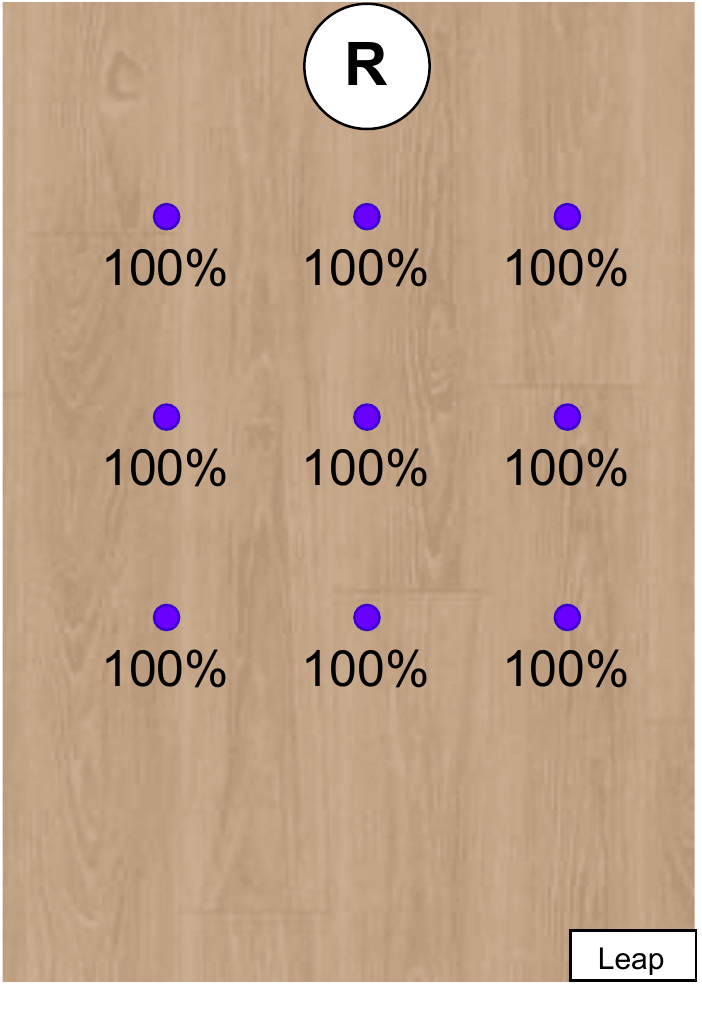}
      \caption{With feedback}
      \label{fig:sub2}
    \end{subfigure}
    \caption{Deictic gesture evaluation results. R represents the robot base, Leap represents the Hand sensor device and blue circles are target objects. The percentage represents the accuracy of both tested users for each object.}
    \label{fig:gesture_evaluation_results}
\end{figure}

Accuracies for detecting the target object in the grid for two users are listed in Tab.~\ref{tab:deictictab1}. Feedback from the simulator (see Fig.~\ref{fig:experimentalsetup}) shows to the user which object is in aim. The better way than observing values on display may be operating the robot over the object that is currently a target. This increases overall system ergonomics and immersion with the system.
More detailed results with accuracies for individual positions are shown in Fig.\ref{fig:gesture_evaluation_results}.

\begin{table}[h]
\caption{Total accuracies of Deictic gesture evaluated on 9 objects grid on 5 users. Firstly, users were given a blind test with no feedback. Secondly, they observed feedback on the screen. Finally, after training the accuracy was evaluated again with no feedback.}
\label{tab:deictictab1}
\begin{center}
  \begin{tabular}{|c|c|l|}
    \hline
    Accuracy [\%] & Group members & Others \\
    \hline
    naive, no feedback   & 59.3 & 40.7   \\
    trained. no feedback & 68.5 & 66.7  \\
    with feedback & 100 & 100 \\
    \hline
  \end{tabular}
\end{center}
\end{table}

\subsection{Gesture classification evaluation}
\label{sec:gestureclassification}

The gesture set needs to be chosen correctly, any two gestures cannot overlay between each other, then the classifier can distinguish between each gesture and correctly classify them. For our sample gesture set, we used 8 static gestures (grab, pinch, point, two, three, four, five, thumbs up), see Tab.~\ref{tab:gestureclassification} for results. 

Dynamic gesture set consists of $5$ gesture including the \textit{no gesture}. The gestures represent directional swipes and are tuned for right-hand usage, the training dataset consists of $\sim300$ demonstrations, and the benchmark (see Tab.~\ref{tab:gestureclassification}) is made on the training dataset because no learning is happening.

\begin{table}[h]
  \caption{Gesture classification benchmarks. Accuracy represents balanced accuracy $BA = \frac{1}{2} \cdot (\frac{TP}{TP + FN} + \frac{TN}{TN + FP})$, where T/F is True/False and P/N is Positive/Negative (more in [link]) and is evaluated on the test dataset. The static category has around 12000 samples and 4000 test samples.}
  \label{tab:gestureclassification}
\begin{center}
  \begin{tabular}{|c|c|l|}
    \hline
    Type & Method & Accuracy \\
    \hline
    \multirow{2}{*}{Static} & Deterministic & 84.3\%  \\
     & Probabilistic NN & 99.1\%  \\
    \hline
    \multirow{2}{*}{Dynamic} & Euler & 72.0\%  \\
     & DTW & 87.3\%  \\
    \hline
  \end{tabular}
  \end{center}
  \vspace{-0.5cm}
\end{table}
\subsection{Scenarios evaluation}

Users were introduced to gestures by which they can command the robot. After users got familiar with execution, they were able to perform low-level tasks (e.g., pick up, pour) with no problems. Most of the errors happened thanks to the object misdetections: e.g., CosyPose didn't recognize objects on the scene, added ghost objects on the scene randomly, or was crashing. Exemplar execution of the scenarios listed in Sec.~\ref{sec:scenariodescription} can be seen in the accompanying video and on the project webpage \href{http://imitrob.ciirc.cvut.cz/gestureSentence.html}{http://imitrob.ciirc.cvut.cz/gestureSentence.html}. 

\begin{figure}[h]
  \centering
  \includegraphics[width=\linewidth]{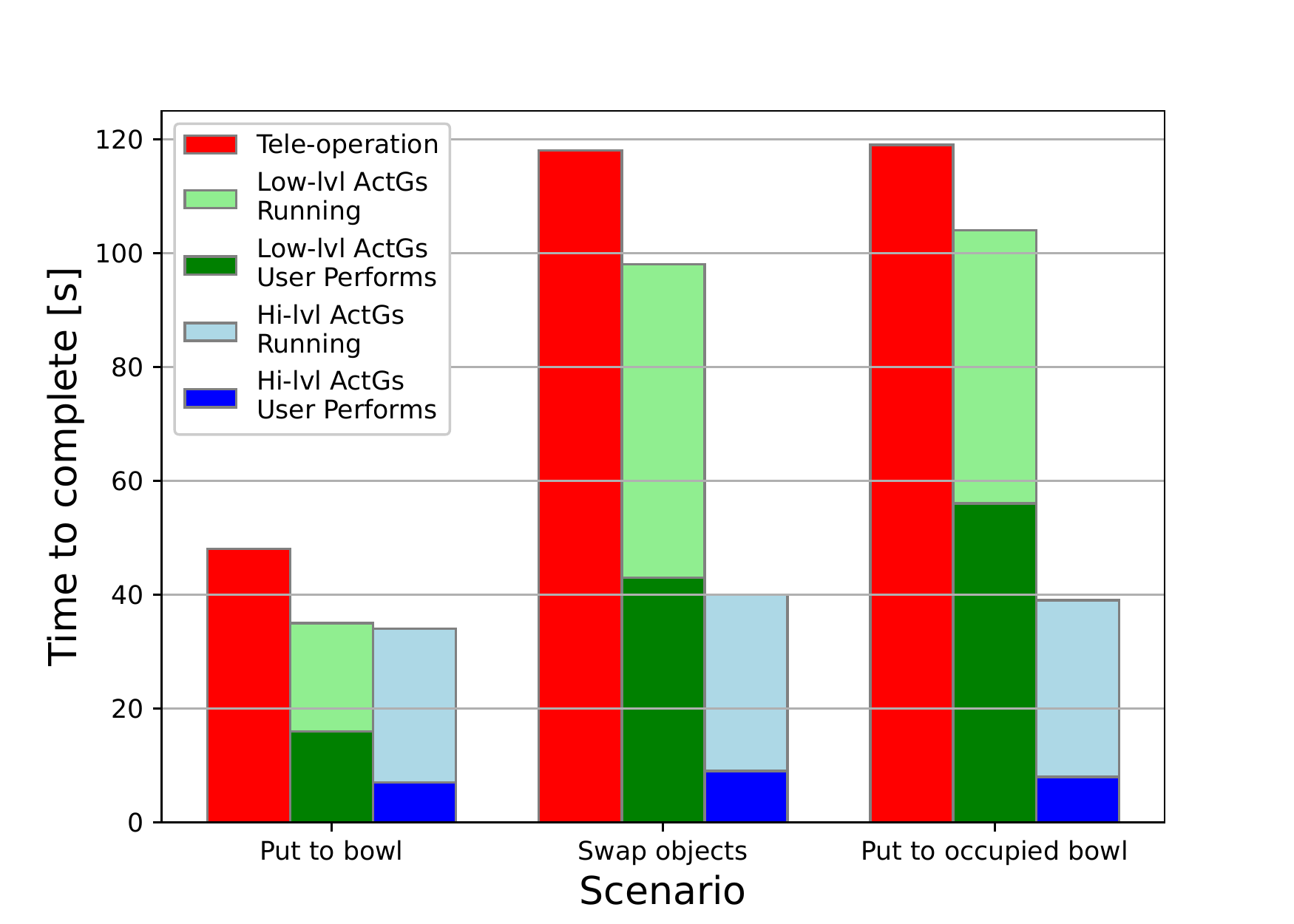}
  \caption{Task completion by time per scenario. Only successful attempts were included. 
  %\jb{check grammar:} There can be seen as a difference between the system running and the time when a user needs to command the system. 
  The system reduced user time needed by around 50\% than total running time. The unfeasible placement task was the hardest to command the robot. Therefore it had the highest failure rate, it can be seen that even higher time commanding the robot.}
  \label{fig:CVWW22_plot_taskcompletion}
\end{figure}

The three selected scenarios (see Sec.~\ref{sec:scenariodescription}) were evaluated more thoroughly. Parameters like time completion, success rate, and percentage of human intervention were noted. The completion time of each scenario when using Teleoperation, Low-level actions and High-level actions is visualized in Fig.~\ref{fig:CVWW22_plot_taskcompletion}. You can see that teleoperating the robot directly needed the highest execution time and the user had to be also properly trained. %For the new naive person the completion time was more than twice as presented. The Low-level mode needs more actions to complete the task. Hi-level mode typically uses only one action gesture to complete the task if it is properly formulated. 
The time needed to fulfil the task was decreasing during individual trials. %goal was decreasing when performing each next task. The input user gesture sentence must be performed correctly to assign trigger the correct intent. When the wrong action is triggered it may bring the scene to a state from which is not easy to continue (e.g., drop the box on a longer side, which cannot be easily grasped by the robot gripper). This was the main failure type while testing scenarios.

%% file: 07_conclusion.tex
We proposed a system that can handle various gesture expressions while handling different gesture types (Action gestures, Deictic, etc.) and combining its information to express and execute the user intent.

Deictic test on users gave us information that visual feedback is a crucial component of such a system.%we always need visual feedback, which guarantees us the correct target aim, because otherwise there may be errors due to bad hand bone structure values, which may be caused by the worse bad seeing angle of the tracked hand.

From the scenario test, we can see that having preprogrammed general, but high-level tasks results in task completion at the lowest time when gestures are performed correctly. Teleoperation is the slowest approach, but it can handle more fine-grained tasks. In the future, it may serve us as teaching the robot new skill. When the skill is remembered, the low-level action gesture may be mapped to this skill. Then the action intent doesn't have to be manually programmed. Controlling the robot via action gestures lowered the execution task by up to 60\%, than using a direct teleoperation method.

%% file: root.bbl
\begin{thebibliography}{10}
\providecommand{\url}[1]{#1}
\csname url@rmstyle\endcsname
\providecommand{\newblock}{\relax}
\providecommand{\bibinfo}[2]{#2}
\providecommand\BIBentrySTDinterwordspacing{\spaceskip=0pt\relax}
\providecommand\BIBentryALTinterwordstretchfactor{4}
\providecommand\BIBentryALTinterwordspacing{\spaceskip=\fontdimen2\font plus
\BIBentryALTinterwordstretchfactor\fontdimen3\font minus
  \fontdimen4\font\relax}
\providecommand\BIBforeignlanguage[2]{{%
\expandafter\ifx\csname l@#1\endcsname\relax
\typeout{** WARNING: IEEEtran.bst: No hyphenation pattern has been}%
\typeout{** loaded for the language `#1'. Using the pattern for}%
\typeout{** the default language instead.}%
\else
\language=\csname l@#1\endcsname
\fi
#2}}

\bibitem{Dragan_Srinivasa}
A.~D. Dragan and S.~S. Srinivasa, \emph{Formalizing assistive
  teleoperation}.\hskip 1em plus 0.5em minus 0.4em\relax MIT Press, July, 2012,
  vol. 376.

\bibitem{10.3115/991365.991471}
\BIBentryALTinterwordspacing
A.~Kobsa, J.~Allgayer, C.~Reddig, N.~Reithinger, D.~Schmauks, K.~Harbusch, and
  W.~Wahlster, ``Combining deictic gestures and natural language for referent
  identification,'' in \emph{Proceedings of the 11th Coference on Computational
  Linguistics}, ser. COLING '86.\hskip 1em plus 0.5em minus 0.4em\relax USA:
  Association for Computational Linguistics, 1986, p. 356–361. [Online].
  Available: \url{https://doi.org/10.3115/991365.991471}
\BIBentrySTDinterwordspacing

\bibitem{Nesnas_Fesq_Volpe_2021}
I.~A. Nesnas, L.~M. Fesq, and R.~A. Volpe, ``\BIBforeignlanguage{en}{Autonomy
  for space robots: Past, present, and future},''
  \emph{\BIBforeignlanguage{en}{Current Robotics Reports}}, vol.~2, no.~3, p.
  251–263, Sep 2021.

\bibitem{Vanc2023}
P.~Vanc, J.~K. Behrens, and K.~Stepanova, ``Context-aware robot control using
  gesture episodes,'' in \emph{2018 IEEE/RSJ ICRA}, 2023.

\bibitem{zhang2019gesture}
W.~Zhang, H.~Cheng, L.~Zhao, L.~Hao, M.~Tao, and C.~Xiang, ``A gesture-based
  teleoperation system for compliant robot motion,'' \emph{Applied Sciences},
  vol.~9, no.~24, p. 5290, 2019.

\bibitem{neto2019gesture}
P.~Neto, M.~Sim{\~a}o, N.~Mendes, and M.~Safeea, ``Gesture-based human-robot
  interaction for human assistance in manufacturing,'' \emph{The International
  Journal of Advanced Manufacturing Technology}, vol. 101, pp. 119--135, 2019.

\bibitem{nuzzi2021meguru}
C.~Nuzzi, S.~Pasinetti, R.~Pagani, S.~Ghidini, M.~Beschi, G.~Coffetti, and
  G.~Sansoni, ``Meguru: a gesture-based robot program builder for
  meta-collaborative workstations,'' \emph{Robotics and Computer-Integrated
  Manufacturing}, vol.~68, p. 102085, 2021.

\bibitem{mazhar2019real}
O.~Mazhar, B.~Navarro, S.~Ramdani, R.~Passama, and A.~Cherubini, ``A real-time
  human-robot interaction framework with robust background invariant hand
  gesture detection,'' \emph{Robotics and Computer-Integrated Manufacturing},
  vol.~60, pp. 34--48, 2019.

\bibitem{behrens2019specifying}
J.~K. Behrens, K.~Stepanova, R.~Lange, and R.~Skoviera, ``Specifying dual-arm
  robot planning problems through natural language and demonstration,''
  \emph{IEEE Robotics and Automation Letters}, vol.~4, no.~3, pp. 2622--2629,
  2019.

\bibitem{losey2018review}
D.~P. Losey, C.~G. McDonald, E.~Battaglia, and M.~K. O'Malley, ``A review of
  intent detection, arbitration, and communication aspects of shared control
  for physical human--robot interaction,'' \emph{Applied Mechanics Reviews},
  vol.~70, no.~1, 2018.

\bibitem{8593766}
S.~Jain and B.~Argall, ``Recursive bayesian human intent recognition in
  shared-control robotics,'' in \emph{2018 IEEE/RSJ International Conference on
  Intelligent Robots and Systems (IROS)}, 2018, pp. 3905--3912.

\bibitem{jonnavittula2022communicating}
A.~Jonnavittula and D.~P. Losey, ``Communicating robot conventions through
  shared autonomy,'' in \emph{2022 International Conference on Robotics and
  Automation (ICRA)}.\hskip 1em plus 0.5em minus 0.4em\relax IEEE, 2022, pp.
  7423--7429.

\bibitem{Colledanchise_Ogren_2018}
\BIBentryALTinterwordspacing
M.~Colledanchise and P.~Ögren, \emph{Behavior Trees in Robotics and
  {AI}}.\hskip 1em plus 0.5em minus 0.4em\relax {CRC} Press, jul 2018.
  [Online]. Available: \url{https://doi.org/10.1201\%2F9780429489105}
\BIBentrySTDinterwordspacing

\bibitem{Weichert_Bachmann_Rudak_Fisseler_2013}
F.~Weichert, D.~Bachmann, B.~Rudak, and D.~Fisseler,
  ``\BIBforeignlanguage{en}{Analysis of the accuracy and robustness of the leap
  motion controller},'' \emph{\BIBforeignlanguage{en}{Sensors}}, vol.~13,
  no.~55, p. 6380–6393, May 2013.

\bibitem{gesturetoolbox}
P.~Vanc, ``{Gesture teleoperation toolbox v.0.1},''
  \url{https://github.com/imitrob/teleop gesture toolbox/}, 2022, [Online;
  accessed 2-March-2023].

\bibitem{Emaasit_2018}
\BIBentryALTinterwordspacing
D.~Emaasit, ``Pymc-learn: Practical probabilistic machine learning in python,''
  no. arXiv:1811.00542, Oct 2018, arXiv:1811.00542 [cs, stat]. [Online].
  Available: \url{http://arxiv.org/abs/1811.00542}
\BIBentrySTDinterwordspacing

\bibitem{Forbes_Fiume_2005}
\BIBentryALTinterwordspacing
K.~Forbes and E.~Fiume, ``\BIBforeignlanguage{en}{An efficient search algorithm
  for motion data using weighted pca},'' in
  \emph{\BIBforeignlanguage{en}{Proceedings of the 2005 ACM
  SIGGRAPH/Eurographics symposium on Computer animation - SCA ’05}}.\hskip
  1em plus 0.5em minus 0.4em\relax Los Angeles, California: ACM Press, 2005,
  p.~67. [Online]. Available:
  \url{http://portal.acm.org/citation.cfm?doid=1073368.1073377}
\BIBentrySTDinterwordspacing

\bibitem{Paraschos_Daniel_Peters_Neumann_2013}
\BIBentryALTinterwordspacing
A.~Paraschos, C.~Daniel, J.~R. Peters, and G.~Neumann, ``Probabilistic movement
  primitives,'' in \emph{Advances in Neural Information Processing Systems},
  vol.~26.\hskip 1em plus 0.5em minus 0.4em\relax Curran Associates, Inc.,
  2013. [Online]. Available:
  \url{https://proceedings.neurips.cc/paper/2013/hash/e53a0a2978c28872a4505bdb51db06dc-Abstract.html}
\BIBentrySTDinterwordspacing

\bibitem{Muller_2007}
\BIBentryALTinterwordspacing
\emph{\BIBforeignlanguage{en}{Dynamic Time Warping}}.\hskip 1em plus 0.5em
  minus 0.4em\relax Berlin, Heidelberg: Springer, 2007, p. 69–84. [Online].
  Available: \url{https://doi.org/10.1007/978-3-540-74048-3_4}
\BIBentrySTDinterwordspacing

\bibitem{murphy2013machine}
\BIBentryALTinterwordspacing
K.~P. Murphy, \emph{Machine learning : a probabilistic perspective}.\hskip 1em
  plus 0.5em minus 0.4em\relax Cambridge, Mass. [u.a.]: MIT Press, 2013.
  [Online]. Available:
  \url{https://www.amazon.com/Machine-Learning-Probabilistic-Perspective-Computation/dp/0262018020/ref=sr_1_2?ie=UTF8\&qid=1336857747\&sr=8-2}
\BIBentrySTDinterwordspacing

\bibitem{Kucukelbir_Tran_Ranganath_Gelman_Blei_2016}
A.~Kucukelbir, D.~Tran, R.~Ranganath, A.~Gelman, and D.~M. Blei, ``Automatic
  differentiation variational inference,'' \emph{Journal of machine learning
  research}, 2017.

\bibitem{Rohmer_Singh_Freese_2013}
E.~Rohmer, S.~P.~N. Singh, and M.~Freese, ``V-rep: A versatile and scalable
  robot simulation framework,'' in \emph{2013 IEEE/RSJ International Conference
  on Intelligent Robots and Systems}, Nov 2013, p. 1321–1326.

\bibitem{James_Freese_Davison_2019}
\BIBentryALTinterwordspacing
S.~James, M.~Freese, and A.~J. Davison, ``Pyrep: Bringing v-rep to deep robot
  learning,'' no. arXiv:1906.11176, Jun 2019, arXiv:1906.11176 [cs]. [Online].
  Available: \url{http://arxiv.org/abs/1906.11176}
\BIBentrySTDinterwordspacing

\bibitem{labbe2020}
Y.~Labb{\'e}, J.~Carpentier, M.~Aubry, and J.~Sivic, ``Cosypose: Consistent
  multi-view multi-object 6d pose estimation,'' in \emph{Computer Vision--ECCV
  2020: 16th European Conference, Glasgow, UK, August 23--28, 2020,
  Proceedings, Part XVII 16}.\hskip 1em plus 0.5em minus 0.4em\relax Springer,
  2020, pp. 574--591.

\bibitem{7254318}
B.~Calli, A.~Walsman, A.~Singh, S.~Srinivasa, P.~Abbeel, and A.~M. Dollar,
  ``Benchmarking in manipulation research: Using the yale-cmu-berkeley object
  and model set,'' \emph{IEEE Robotics \& Automation Magazine}, vol.~22, no.~3,
  pp. 36--52, 2015.

\end{thebibliography}
